\documentclass[preprint,12pt]{ elsarticle}

\usepackage{amssymb}
\usepackage{float}
\restylefloat{table}
\usepackage{xcolor,soul}
\usepackage{amsmath}
\usepackage{algorithm}
\usepackage{algpseudocode}

\biboptions{numbers,sort&compress}
\journal{Applied Mathematical Modelling}

\begin{document}

\begin{frontmatter}



\title{Identification of release sources in advection-diffusion system by machine learning combined with Green's function inverse method}


\author{Valentin G. Stanev}
\address{Physics and Chemistry of Materials Group\\
	Theoretical Division\\
	Los Alamos National Laboratory\\
	Los Alamos, NM, USA} 

\author{Filip L. Iliev}
\address{Physics and Chemistry of Materials Group\\
	Theoretical Division\\
	Los Alamos National Laboratory\\
	Los Alamos, NM, USA} 

\author{Scott Hansen}
\address{Computational Earth Science Group\\
	Earth and Environmental Sciences Division\\ 
	Los Alamos National Laboratory\\
	Los Alamos, NM, USA}

\author{Velimir V.
	Vesselinov}
\address{Computational Earth Science Group\\
	Earth and Environmental Sciences Division\\ 
	Los Alamos National Laboratory\\
	Los Alamos, NM, USA}

\author{Boian S.
	Alexandrov\corref{cor1}}
\ead{boian@lanl.gov}
\cortext[cor1]{Corresponding author}
\address{Physics and Chemistry of Materials Group\\
	Theoretical Division\\
	Los Alamos National Laboratory\\
	Los Alamos, NM, USA}

\begin{abstract}
The identification of sources of advection-diffusion transport is 
 based usually on solving complex ill-posed inverse models against the available state-variable data records. However, if there are several sources with different locations and strengths, the data records represent mixtures rather than the separate influences of the original sources. Importantly, the number of these original release sources is typically unknown, which hinders reliability of the classical inverse-model analyses. To address this challenge, we present here a novel hybrid method for identification of the unknown number of release sources. Our hybrid method, called HNMF, couples unsupervised learning based on Non-negative Matrix Factorization (NMF) and inverse-analysis Green's functions method. HNMF synergistically performs decomposition of the recorded mixtures, finds the number of the unknown sources and uses the Green's function of advection-diffusion equation to identify their characteristics. In the paper, we introduce the method and demonstrate that it is capable of identifying the advection velocity and dispersivity of the medium as well as the unknown number, locations, and properties of various sets of synthetic release sources with different space and time dependencies, based only on the recorded data. HNMF can be applied directly to any problem controlled by a partial-differential parabolic equation where mixtures of an unknown number of sources are measured at multiple locations.

\end{abstract}

\begin{keyword}
Advection-diffusion transport;
Inverse problem;
Source identification; 
Non-negative matrix factorization;
Green functions
\end{keyword}

\end{frontmatter}

\section{Introduction}
\label{Sec1}
In the last several decades one of the most important research topics in the hydrogeological sciences has been the contaminant transport in subsurface environment \cite{gelhar1993stochastic, fetter1999contaminant}.
The research has been driven by substantial challenges associated with prediction and remediation of contaminant plumes in the environment.
Most of these challenges are due to uncertainties associated with contaminant sources (e.g. source locations, emission strengths, release transients, etc.) as well as contaminant migration (e.g., velocity, dispersivity, etc., related to aquifer and contaminant transport properties).
Determining the number of the contaminant sources, their locations and physical properties, is an important task that yields valuable information needed for prediction of the fate and transport of contaminant, the hazard and risk assessments, and remediation.
Most often, the information about the contamination sources and contaminant migration in the medium (an aquifer) is limited or not available, which explains the increasing use of complex numerical inverse model analyses \cite{guan2006identification, atmadja2001pollution,Sakthivel2011571, borukhov2015identification, mamonov2013point,hamdi2013inverse, murray2014spatio} and multivariate statistical and machine learning techniques \cite{chan2003artificial, khalil2005applicability, rasekh2012machine, manca2013case, shahraiyni2015new}. These methods address the need for accurate predictions of perpetually increasing number of environmental management problems caused by contamination groundwater-supply resources  \cite{vengosh2014critical}.

The existing methods for contaminant source identification rely on the available contamination site observations.
The tools for observation of contaminant plume are various types of detector (sensor) arrays recording the spatiotemporal behavior of the contaminant plumes.
The detectors are typically monitoring wells which are sampled with some temporal regularity.

Importantly, these detectors typically measure mixtures of signals originating from  an unknown number of contaminant sources, which presents a considerable challenge to modeling.
The mixing ratios of the different sources at each sensor are also usually unknown. All these uncertainties hamper the reliability of the standard inverse-model analyses. An alternative approach is to use model-free Blind Source Separation (BSS) methods, commonly applied to signal processing problems. The BSS methods, however, do not exploit the available understanding of the physics of the contaminant processes. 

To address this problem we have developed and present here a new hybrid inverse method, which we call HNMF, capable of identification of the unknown number and physical properties of release sources by combining inverse methods and machine learning algorithms. HNMF is utilizing (a) the Green's function of the corresponding partial-differential equation that governs the physics of the monitored process (advection-diffusion equation in the case of contaminant transport) and (b) a  BSS method \cite{belouchrani1997blind}, based on Non-negative Matrix Factorization (NMF) \cite{lee1999learning}, combined with a custom-made semi-supervised clustering algorithm, used to un-mix the observations and identify the release sources.

To validate HNMF we generate several synthetic datasets, representing various problem settings. We generated four sets with different number of point-like instantaneous sources in infinite medium, recorded by different number of detectors, and three sets of sources with more complicated space and time dependences, including sources with a constant release rate, and sources with finite sizes. We also consider processes based on point-like instantaneous sources in a medium bounded by a reflecting boundary.
For all these cases, HNMF accurately determines the number, locations, and physical characteristics of the unknown contaminant sources from observed samples of their mixtures, without any additional information about the sources or the physical properties of the medium where the contaminant transport occurs.
The method also estimates the transport properties of the medium, such as the advective velocity and dispersivity. Furthermore, HNMF can be also applied with small or no changes to many typical problems involving heat transport (see, e.g., Refs. \cite{Lin20072696,Lin20112607}).

By combining model-free machine learning BSS method with inverse-problem analysis we are able to solve problems, which present a significant challenge for usual BSS methods and to classical inverse methods, but are very common in practical applications: an unknown number of sources spreading signals that change their form as they propagate, either because of the fundamental nature of the propagation process, or because of the properties of the medium.

\section{Statement of the problem}
\label{Sec2}
The main goal of the paper is to present a novel hybrid method, which we call HNMF, that combines the Green's function of the advection-diffusion equation with a semi-supervised adaptive machine learning approach for identification of contaminant sources in porous media, based on a set of observations.
We assume that the observations are taken at several detectors, dispersed in space, and monitoring contaminant transients over a period of time.
If there are multiple contamination sources in the aquifer, each detector registers a mixture of contamination fields (plumes) originating from sources  at different release locations.
It is assumed that each contaminant source is releasing the same geochemical constituent that is mixed in the aquifer and the resultant mixture is detected at the observation points (detectors).
If each source was releasing different geochemical constituents similar to our methodology can be applied with minor adjustments.
We also assume that the geochemical constituent are conservative (non-reactive) and their transport is not impacted by geochemical reactions or other fluid/solid interactions in the porous media where the flow occurs.
Our objective is to identify the unknown number, release locations and physical parameters of the original contamination sources. This objective requires first to decompose the recorded mixtures to their constituent components.
This decomposition allows us to use the Green's function of the advection-diffusion equation to extract the physical parameters, location coordinates and time dependence of the sources.
In the rest of this section, we present the general mathematical formulation of the problem and introduce the notations we are going to use. 

\subsection{Advection-diffusion equation}

The transport of solute in a medium is described by linear partial-differential parabolic equation \cite{bear2013dynamics}:
\begin{eqnarray}\label{gen_eq}
\frac{\partial C}{\partial t} = \triangledown({\bf D} \cdot \triangledown C) - \triangledown \cdot ({\bf u} C) + L C + Q.
\end{eqnarray}
This equation describes the rate of change of the concentration of the solute/contaminant $C({\bf x}, t)$, defined in some (space and time) domains: ${\bf x} \in \mathbb{R}^d$ and $t \in [t_{init}, t_{final}]$.
The matrix ${\bf D}$ is the hydrodynamic dispersion, which is a combination of molecular diffusion and mechanical dispersion (in porous media the latter typically dominates), and ${\bf u}$  is the advective velocity.
$Q$ is a source function, representing possible sinks and sources of contaminants.
The term involving $L$ describes possible chemical transformations of the contaminant.
In the following, we will neglect this term, assuming that $L = 0$.
This type of equation also describes heat transport in various media \cite{li2014heat}, and the procedure presented here can be applied to this problem as well.

For simplicity, we will consider a (quasi-)two-dimensional medium, so ${\bf x}\equiv (x,y) \in \mathbb{R}^2$; we assume that the third dimension $z$ is a constant, $z=z_0$, and that the extend along $z$ of the aquifer is small, and thus $C$ is uniformly distributed in that direction.
By assuming that both ${\bf u}$ and ${\bf D}$ are independent of ${\bf x}$ and choosing the $x$-axis along the direction of ${\bf u}$,  (\ref{gen_eq}) becomes:
\begin{eqnarray}\label{gen_eq2}
\frac{\partial C}{\partial t} = \triangledown({\bf D }\cdot \triangledown C) - \triangledown \cdot ({\bf u} C) + Q \equiv D_x \frac{\partial^2 C}{\partial x^2} + D_y \frac{\partial^2 C}{\partial y^2} - u_x \frac{\partial C}{\partial x} + Q.
\end{eqnarray}
Note that $D_x \neq D_y$ because the advection motion causing mechanical dispersion breaks the isotropy of space.

To uniquely specify a solution of (\ref{gen_eq2}), we need to impose initial and boundary conditions.
We assume that there is no contamination before the sources start emitting, and the initial condition is $C(t<$ min$(t_s)) = 0$.
 Different types of boundary conditions (Dirichlet, Neumann or Cauchy) can be appropriate, depending on the type of the boundary itself (inflow, outflow or impenetrable).
For simplicity, we will focus mainly on the case of an infinite two-dimensional space, assuming that the aquifer is large enough so its boundaries do not affect the distribution of $C$ (at least at the time-scales of interest).
In this case, the boundary condition is given by $C \rightarrow 0 $ as ${\bf x}\rightarrow \infty$. 

Since (\ref{gen_eq2}) is a linear partial differential equation, we can \st{use the principle of superposition to} write the solution with specified source term:
\begin{eqnarray}\label{eq:Int}
C(x, t) = \int dt' d{\bf x}' G({\bf x} - {\bf x}',t - t') Q({\bf x}', t'),
\end{eqnarray} 
where $G({\bf x}, t)$ is the Green's function of the diffusion-advection equation, and the sources are described by $ Q({\bf x}, t)$. The Green's function is solution of (\ref{gen_eq2}) with (space and time) point source:
\begin{eqnarray}\label{eq:pde}
\frac{\partial G}{\partial t} = D_x \frac{\partial^2 G}{\partial x^2} + D_y \frac{\partial^2 G}{\partial y^2} - u_x \frac{\partial G}{\partial x} + \delta( x ) \delta( y ) \delta(t).
\end{eqnarray}
Here, the source is located at ${\bf x}_s=(0,0)$, it is active at $t_s=0$, and $\delta(...)$ denotes the Dirac delta-function.
The solution of (\ref{eq:pde}) is well known \cite{park2001analytical}, and can be written as:
\begin{eqnarray}\label{eq:Green}
G({\bf x},t) = \frac{1}{4 \pi \sqrt{D_x D_y} t} e^{-\frac{(x - u_x t)^2}{4 D_x t}}e^{-\frac{y ^2}{4 D_y t}},
\end{eqnarray}

In the case $Q$ represents a collection of localized in space and time sources, it can be written as $Q \sim \sum q_s \delta(x-x_s) \delta(y-y_s) \delta(t-t_s)$, where  $s$ indexes the contamination sources, $x_s$, $y_s$ and $t_s$ refer to the coordinates and release time of the $s$-th  source and $q_{s}$ specifies its strength. Then the integral in (\ref{eq:Int}) can be  replaced with a sum over $s$: 
\begin{eqnarray}
C({\bf x}, t) = \sum_{s=1}^{N_{s}} q_{s} G({\bf x}- {\bf x}_s,t - t_s),
\end{eqnarray}
where $N_{s}$ is the total number of sources.

Despite the fact that so far we have considered explicitly only point-like instantaneous sources, the method can be straightforwardly applied to sources with more complicated space and time dependency: $Q =  \sum_{s=1}^{N_{s}}{q_{s}({\bf x}, t)}$.
In this case the solution of (\ref{gen_eq2}) can be written as: 
\begin{eqnarray}\label{eq:Green_int}
C({\bf x}, t) = \sum_{s=1}^{N_{s}} \int d{\bf x'} dt' G({\bf x}- {\bf x'},t - t') q_{s}({\bf x'}, t'),
\end{eqnarray}
where $q_{s}({\bf x},t)$ now encodes the extend (in space and time) of the $s$-th source. 
If the above integral (with known functional dependences $q_{s}({\bf x'}, t')$) can be solved analytically, the result for $C({\bf x}, t)$ reduces to a closed-form expression (cf. \cite{park2001analytical}).
However, even if this is not the case, the method we present here works for $C({\bf x}, t)$ obtained by using a numerical integration. 

\subsection{Blind Source Separation (BSS) methods}

In a typical problem that requires BSS methods \cite{belouchrani1997blind}, the observed data matrix, ${V}$, (${V}\in{M}_{N;T}({\mathbb{R}})$), is formed by a linear mixing (at each one of the $N$ detectors) of $N_s$ unknown original signals presented in the sources' matrix, $ {H}$, ($ {H}\in {M}_{N_s;T}({\mathbb{R}})$), blended by an also unknown mixing matrix, $ {W}$, ($ {W}\in {M}_{N;N_s}({\mathbb{R}})$), where $\mathbb{R}$ denotes the set of real numbers. Thus, at a given detector $n$ at a moment of time $t$, we can write:
\begin{equation}\label{eq:BSS}
{V}_{n}(t)= \sum _{s}{W}_{n,s} {H}_{s}(t) + {\epsilon_{n}(t)},
\end{equation}
where ${\epsilon_n(t)}\in {M}_{N;T}(\mathbb{R})$, and denotes presence of possible noise or unbiased errors in the measurements (also unknown).
The problem is usually solved in a temporally discretized framework, and the goal of a BSS method is to identify the $N_s$ original sources.
Since both factors ${H}$ and ${W}$ are unknown, and even their size $N_s$ (i.e., the number of sources mixed at each detector record) is unknown, the main difficulty in solving this problem is that it is under-determined.

When ${V}$ is a non-negative matrix, one of the most widely-used BSS methods is NMF -- an unsupervised learning method, created for parts-based representation \cite{fischler1973representation} in the field of image recognition \cite{paatero1994positive, lee1999learning}, that was successfully leveraged for decomposition of mixtures formed by various types of signals \cite{cichocki2009Non-negative}.
NMF enforces a non-negativity constraint on both the original signals in ${H}$ and their mixing components in ${W}$, and can successfully decompose large sets of non-negative observations, ${V}$, by leveraging, for example, the multiplicative update algorithm \cite{lee1999learning}. However, the NMF algorithm requires \emph{prior} knowledge of the number of the original sources.

\subsection{The HNMF Method}

If we wanted to identify signals originated from a number of known sources, which mixtures are recorded by set of detectors, then the explicit form of the Green's function, (\ref{eq:Green}), could be enough to solve this problem.
However, here we are aiming to solve a more complex inverse problem: to determine the characteristics of an \emph{unknown number} of sources based on the mixtures of signals recorded by multiple detectors positioned at arbitrary locations.
Apparently this is a problem that requires a BSS method, and because the signals are non-negative, NMF appears to be a good match for the task.
However, despite their advantages, none of the conventional BSS methods are directly appropriate for this task.
The issue is in the nature of the process of the advection-diffusion transport itself.

Indeed, while some physical processes (e.g., processes subject to dispersionless wave equation) permit transport in which the signals keep their form undistorted as they travel through the medium, the parabolic equation describes diffusion combined with advection.
To examine the consequences of this, let us consider a single point source.
Detectors situated at different distances from the source will record signals, differing in shape and time dependence.
This distortion is difficult to be treated by the model-free BSS methods that do not exploit the available scientific understanding of the advection-diffusion transport.

The hybrid method we are proposing here takes advantage of the knowledge of the physical processes involved in the advection-diffusion transport, by using the explicit form of the Green's function of the advection-diffusion equation, and estimates the ({\it a priori} unknown) number of release sources, based on the solutions' robustness.

\section{Methodology}

The method we propose in this work has two well-separated stages: (a) Non-negative Matrix Factorization based on nonlinear minimization and explicit type of the Green function and (b) custom semi-supervised clustering algorithm, leveraged to estimate the unknown number of sources based on the robustness of the solutions. 

\subsection{Non-negative Matrix Factorization}
Based on the Green's function of advection-diffusion equation, we can use the explicit form of the original signals at times $t$ ($t=1,2,..,T$) and at the locations of each of the $N$ detectors: ($x_n$, $y_n$), $n=1,2,..,N$.
These signals originate from sources located at points with coordinates: ($x_s$, $y_s$), $s=1,2,...,N_s$, with source strengths $q_s$.
Therefore, we have to solve a NMF-type of minimization:
\begin{equation}\label{eq:BSScontaminant}
{V}_{n} (t)= \sum^{N_s} _{ s=1}{W}_{s} {H}_{s,n}(t) + {\epsilon_n{(t)}}; t=1,2,..,T
\end{equation}
where,
\begin{eqnarray}
{W}_{s}& \equiv & {q}_{s}\geqslant 0,\\
{H}_{s,p}(t)& =& \frac{1}{4 \pi \sqrt{D_x D_y} t} e^{-\frac{((x_{s}- x_{n})- u_x t)^2}{4 D_x t}}e^{-\frac{(y_{s} - y_{s})^2}{4 D_y t}} \geqslant 0.
\end{eqnarray}
and ${\epsilon}$ is the Gaussian noise or unbiased measurement errors.
As explained earlier, the record at each detector is a superposition of the contributions from all $N_s$ sources.
The coordinates of each of the detectors, ($x_n$, $y_n$) and the functional form of the Green function, $G_s$ are assumed known, while the parameters $q_{s}$, $x_{s}$ and $y_{s}$, as well as the contaminant transport characteristics $u_x$, $D_x,$ and $D_y$ are the unknown parameters.
The goal of the minimization procedure is to obtain the physical parameters and transport characteristics that reproduce (with some accuracy) the observed data, rather than to reconstruct an entirely unknown functional dependence.
This considerably simplifies the problem, and for many cases, the minimization can be carried by well-known nonlinear least-squares procedures (for example, Levenberg-Marquardt algorithm) as implemented in the open source code MADS (http//mads.lanl.gov; \cite{MadsJulia2016, Vesselinov2016AGUa}) and applied to minimize the cost function $O$,
\begin{equation}
\begin{aligned} 
\label{eq:NMFof}
{O} = \sum_{n=1}^N\sum_{t=1}^{T} \left(V_{n, t}-\sum_{s=1}^{N_s} W_{s} H_{s,n,t} \right)^2\\
\text{where, } W_{s} \geqslant 0; H_{s,n,t} \geqslant 0; 
\end{aligned} 
\end{equation}
The minimization of the $l2$ cost function assumes that each measurement at a given space-time point is an independent Gaussian-distributed random variable, which corresponds to the white noise, ${\epsilon}$.
If each detector has its own and different (possibly time-dependent) measurement error, minimizing the cost function in (\ref{eq:NMFof}) should be replaced by a weighted least-squares procedure, where the cost at each point is weighted by the inverse square of its measurement error.
Note that since Gaussian functions form a basis in the functional space, none of them can be presented as a linear combination of the others, which guarantees a uniqueness of the minimum of the cost function in (\ref{eq:NMFof}) \cite{yarotsky2013univariate}.

However, the main problem is that the unknown number of contamination sources $N_s$ cannot be extracted directly by minimizing $O$ in (\ref{eq:NMFof}). Indeed, this number has to be already known in order to build the explicit form of the solution.

\subsection{Clustering algorithm}
If the number of sources $N_s$ is known, the method described in the previous section would be all that is needed: from the best solution of the minimization procedure (with fixed $N_s$), we would extract the desired estimates of the physical parameters, and thus solve the inverse problem.
Unfortunately, the true number of sources is typically unknown and we have to estimate it solely from the observations.
Moreover, the solution of (\ref{eq:NMFof}), is based on some (often highly inaccurate) initial guess for the unknown parameters.
A naive approach would be to explore all possible solutions applying the nonlinear minimization described in the previous section for an entire range of a possible number of sources, and then use the solution with the smallest reconstruction error to estimate $N_{s}$.
However, this is obviously a flawed approach -- over-fitting will certainly lead to over-estimation of the number of sources; more free parameters will generally lead to a better fit, irrespective of how close the estimated number of sources is to the real one.
To determine the optimal value of the unknown number of sources from the observations, we use a custom semi-supervised clustering algorithm.

The original NMF algorithm also requires \emph{prior} knowledge of the number of the original sources. To deal with this indeterminacy different extensions of the method have been proposed \cite{BayesianNMF,BayesianNMF2,BayesianNMF3}.
Most of them use Bayesian model selection framework (see, e.g., Ref. \cite{BayesianNMF}) that requires running computationally intensive Monte Carlo simulations. 
However, recently it was demonstrated that by coupling NMF with a k-means clustering the number of the original sources can be estimated based on the reproducibility of the solutions.
This approach was introduced to decompose the largest available dataset of human cancer genomes \cite{alexandrov2013signatures}, and then extended for decomposition of physical pressure transients \cite{alexandrov2014blind}.
Although heuristic in nature, this method has proven to be accurate and reliable; it also has an important practical advantage that it is relatively easy to implement and use.   

Here we utilize a similar protocol, based on a custom semi-supervised clustering algorithm, to determine the optimal number of sources.
Specifically, our protocol explores consecutively all possible numbers of original sources, $i = 1, 2..., N$ (the number of sources  is assumed to be less than the number of detectors  -- $N_s \leq N$), and then estimates the robustness as well as the accuracy of the minimization of solutions with different $i$.
As a first step, HNMF performs $N$ sets of $M$ minimizations, called runs, where each minimization in a given run is using the same number for the number of sources, with random initial guesses for the unknown parameters.
Each run results in a set $U_i$, containing $M$ solutions of the minimization. If we denote the derived in $p^{th}$ minimization coordinates and amplitudes of the $i$ sources $(x^p_i, y^p_i, q^p_i)$ by $(x, y, q)^p_i$; and  the corresponding set of advection velocities and hydrodynamic dispersion components,  $({u_x^p}_i, {D_x^p}_i, {D_y^p}_i)$, by $(u_x, D_x,D_y)^p_i$, for the set of solutions, ${ U }_{ i }$, of the whole run of minimizations with $i$ sources, we will have,   
\begin{equation}
{ U }_{ i }=([(x, y, q)^1_i, (u_x, D_x,D_y)^1_i],...,[(x, y, q)^M_i, (u_x, D_x,D_y)^M_i])
\end{equation}
where $M$ is the number of minimizations.
Next, HNMF leverages a semi-supervised clustering to assign each of the components of the $M$ resultant $3*i$-component vectors, $ \{h^i_1 =(x, y, q)^1_i,..., h^M_i = (x, y, q)^M_i\}$ to one of the $i$ clusters containing the coordinates and amplitudes of the sources.
This algorithm is a $k$-means clustering with the additional constrain that keeps the number of solutions in each cluster equal.
For example, for the case of $i = 2$, after a run with $M=1,000$ simulations (performed with random initial guesses for the unknown parameters), each of the two clusters will contain $1,000$ solutions.
Note that we have to enforce the condition for the clusters to contain an equal number of solutions since each simulation contributes exactly one solution for each physical parameter.
During clustering the similarity between two elements $h^{i}_1$ and $h^i_2$ is measured using the cosine distance (also known as cosine similarity) \cite{pang2006introduction}, $\rho(h^i_{1},h^i_{2})$,
\begin{equation}
\rho(h^i_{1},h^i_{2}) = 1 - \frac { \sum _{ i=1 }^{ i }{ h^i_{1, i }h^i_{2, i } } }{ \sqrt {   \sum _{ i=1 }^{ i } {  ({h^i_{1, i }})^2 }   }\sqrt {   \sum _{ i=1 }^{ i } {  ({h^i_{2, i }})^2 }   } }.
\label{eq:rho}
\end{equation} 
The main idea for estimating the unknown number of sources is to use the separation between the clusters as a measure of how good is a particular choice of $i$.
The intuitive reasoning behind it is as follows. 
In the case of the underfitting, i.e., for solutions with $i$ less than the actual number of sources, the clustering could be good; for example, several of the sources could be combined to produce a "super-cluster''.
However, the clustering will break down significantly in the case of over-fitting, when $i$ exceeds $N_{s}$.
Indeed, in this case (even if the average reconstruction error of the solution is small), we do not expect the solutions to be well clustered since there is no unique way to reconstruct the solutions with $i > N_s$, and at least some of the clusters are artificial, rather than real entities.
Thus, the separability of the clusters can be used as an indicator of the solution robustness and applied to identify the number of sources.

To quantify this behavior, we utilize Silhouette width of the clusters \cite{rousseeuw1987Silhouettes}, which measures how well the solutions are clustered by comparing the average distance within a cluster with the average distance to the next closest cluster. 
After the clustering, we compute the Silhouette widths of the clusters and construct their average Silhouette width.
This average Silhouette width is a quantitative measure of how separable, and thus how  reproducible are the average solutions (the centroids of the clusters) for each $i$.

In addition to the average Silhouette width, we use the reconstruction error from (\ref{eq:NMFof}) to evaluate the accuracy of the average solutions.
In general, the solution accuracy increases (while the solution robustness decreases) with the increase of the number of unknown sources.
Thus, the average Silhouette width and the average reconstruction error for each of the $i$ cluster solutions can be used to estimate the optimal number of contaminant sources, $N_s$.
Specifically, we select $N_s$ to be equal to the number of sources that accurately reconstruct the observations (i.e., their average reconstruction error is small enough) while clustering of solutions is sufficiently robust (i.e., the average Silhouette width is close to $1$).

An automatic selection procedure can be also formulated with the help of the Akaike Information Criterion ($AIC$) \cite{akaike2011akaike}.
We compare the models with different number of sources $i$ by calculating (for each $i$ separately) the value:
\begin{eqnarray}
AIC = 2 p - 2 \ln (L) = 2[i(3+N)] + N_t \ln \left(\frac{O^{(i)}}{N_t} \right) \
\end{eqnarray}
here (as usual) we disregarded terms  that do not dependent on $i$ and thus are irrelevant.  $L^{(i)}$ is the likelihood functions of the model with given $i$, and we define it to be the average reconstruction error $O^{(i)}$ of the solutions we have kept in the clustering procedure for this particular $i$: $\ln(L^{(i)}) = - (N_t/2)\ln (O^{(i)}/N_t)$ ($N_t=N*T$, is the total number of data points; the product of the number of detectors $N$ by the number of time observations $T$).
The degrees of freedom of a model with $i$ sources is given by $p= i(3+N) +3$, because each new source comes with $3+N$ parameters ($2$ spacial coordinates plus the source strength, as well as $N$ mixing coefficients), in addition to the $3$ parameters characterizing the medium (because their number is fixed they can be discarded just like the other $i$-independent $AIC$ terms).
The $AIC$ is a standard measure of the relative quality of statistical models, which takes into account both the likelihood function  and the degrees of freedom needed to achieve this level of likelihood. 
Choosing the model that minimizes $AIC$ helps avoid over-fitting. It is also important to note that the $AIC$ value depends on the clustering procedure since we calculate $L^{(i)}$ using the physical parameters determined by the centroids of the clusters.   

Based on the above approach, we have developed the HNMF algorithm for identifying the number and characteristics of contaminant sources subject to advection-diffusion equation. 

\section{Results}

\subsection{The pseudo--code of the proposed HNMF Algorithm }
The following pseudo-code implements the proposed HNMF algorithm as described above
\label{sec:algorithm}
\renewcommand{\thealgorithm}{}
\begin{algorithm}[H]
\caption{HNMF pseudo--code}\label{alg:MyAlgorithm}
\begin{algorithmic}[1]

\State{Get the observation matrix $X$ }
\State{Set the maximum number of unknown sources, $N$}
\State{Set the maximum number minimizations for given number of sources, $M$}
\For{\texttt{i = 1 to N}}
 \State{Create empty list $U_{i}$ (eq. 13)}
   	\For{\texttt{j = 1 to M}}
	\State{Generate random initial guesses for [\textbf{x}, q, \textbf{u}, \textbf{D}] and corresponding \phantom . \phantom . \phantom . \phantom . \phantom . $W$ and $H$ (eq. 10 and 11)}
	\State{Minimize $O^j_i$ (eq. 12) via a nonlinear least-squares solver and find \phantom . \phantom . \phantom . \phantom . \phantom . solution for [\textbf{x}, q, \textbf{u}, \textbf{D}]$^j_i$}
	\State{Append the solution to $U_i $}
   	\EndFor
 \State{Sort the elements of $U_i$: [\textbf{x}, q, \textbf{u}, \textbf{D}]$^j_i$, $ j = 1, ..., M$, by the values of  \phantom . \phantom . \phantom . \phantom . reconstruction of $O^j_i$, and remove the worst 10\%}
\State{Do $k$-means clustering for all elements, [\textbf{x}, q, \textbf{u}, \textbf{D}]$^j_i$, $ j = 1, ..., M$, of \phantom . \phantom . \phantom . $U_i$ with $i$ clusters}
\State{Calculate Silhouette values, $S_i$, and clusters' centroids [\textbf{x}, q, \textbf{u}, \textbf{D}]$_i^{cent}$}
\State{Calculate the $O_i^{cent}$ (eq. 12) as a function of the centroids}
\EndFor
\For{\texttt{i = 1 to N}}
\State{Calculate AIC($i$, $O_i^{cent}$) (eq. 15) for average Silhouettes: $S_i > 0.7$}
\EndFor
\end{algorithmic}
\end{algorithm}

\subsection{Synthetic Data}

\begin{figure} 
\centering
\includegraphics[width=5in]{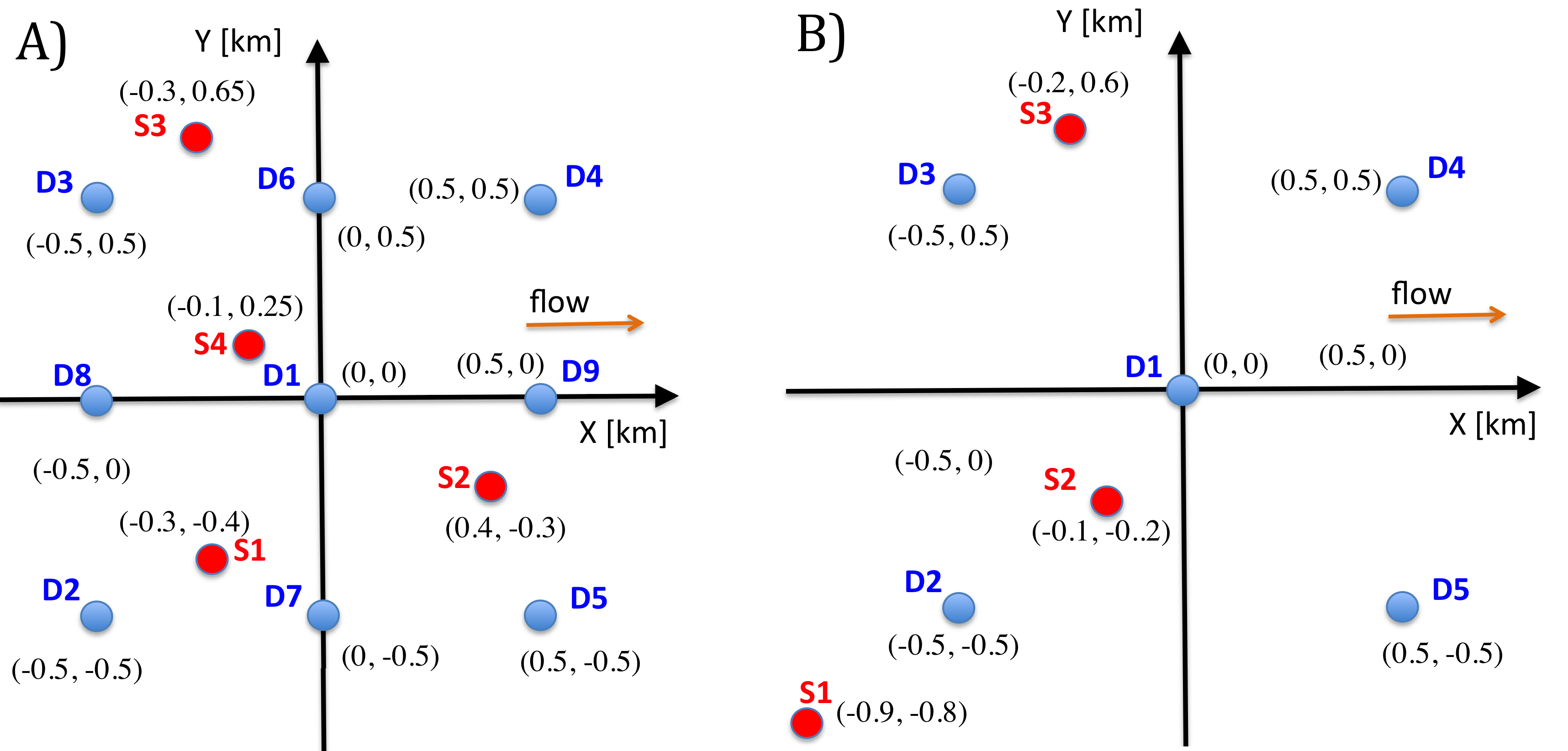}
\caption{Test setups representing arrays of detectors (blue dots) and sources (red dots).
	On panel (A), there are nine detectors and four sources.
	On panel (B), there are five detectors and three sources.}
\label{fig:maps}
\end{figure}

To illustrate HNMF method, we apply it to identify sources from seven distinct  synthetic data sets. We consider two general detector/source configurations as represented in Fig.\ref{fig:maps}.  To make the problem more realistic, we set the physical parameters of the sources and the medium to be in ranges consistent with these observed at actual groundwater contamination sites.
In both setups, the distances are measured in kilometers, the advection velocity is $u = 0.05$ km/year, and its direction was chosen to be along the $x$-axis of the coordinate system.
We have assumed a significant anisotropy in hydrological dispersion: $D_x = 0.005$ km/year$^2$ and $D_y = 0.00125$ km/year$^2$ (thus, $D_y/D_x = 0.25$, due to the presence of advection and mechanical dispersion).
We also assume $t_0$ (the time of source activation) for all sources to be: $t_0=-10$ years ($10$ years before the start of data collection).
We consider a time span of $20$ years (starting with $t=0$), with measurements taken four times per year, so there is a total of $80$ time points for each detector.
By mixing the sources at each detector with added Gaussian noise with strength $10^{-3}$, we construct seven different observation sets (observational matrices) based on the two test configurations presented in Fig.\ref{fig:maps}. For the first configuration the source coordinates are: $S1 = (-0.3, -0.4)$, $S2 = (0.4, -0.3)$, $S3 = (-0.3, 0.65)$, and $S4 = (-0.1, 0.25)$, and all sources are releasing contaminants with constant concentrations equal to $0.5$ mg/L. In the second configuration (Fig.\ref{fig:maps}B), the source coordinates are: $S1 = (-0.9, -0.8)$, $S2 = (-0.1, -0.2)$, and $S3 = (-0.2, 0.6)$, with release concentrations: $0.5$ mg/L, $0.7$ mg/L, and $0.3$ mg/L, respectively.

Starting with random values for the unknown parameters, the minimization procedure runs by adjusting the values of the unknown parameters until the $l2$-norm of the cost function stops changing appreciably, or the maximum number of iterations is reached.
For each possible number of original sources ($i \leq N$), using the constructed observation matrices, we performed runs, each with $M$ simulations (typically  $M \leq 10,000$).
Then, following the algorithm in Section \ref{sec:algorithm}, we gradually pruned the simulations, guided by the quality of clustering. 

\subsection{HNMF Estimates}
We first demonstrate the ability of HNMF to determine the unknown number of point-like instantaneous sources and their parameters in different configurations and in an infinite domain. Next, we fix the number of sources (while keeping this number unknown in the algorithm) and configuration of the detectors and consider sources with different properties and boundary conditions.

\subsubsection{Instantaneous point sources in an infinite domain}

\begin{figure}
\centering\includegraphics[width=5.5in]{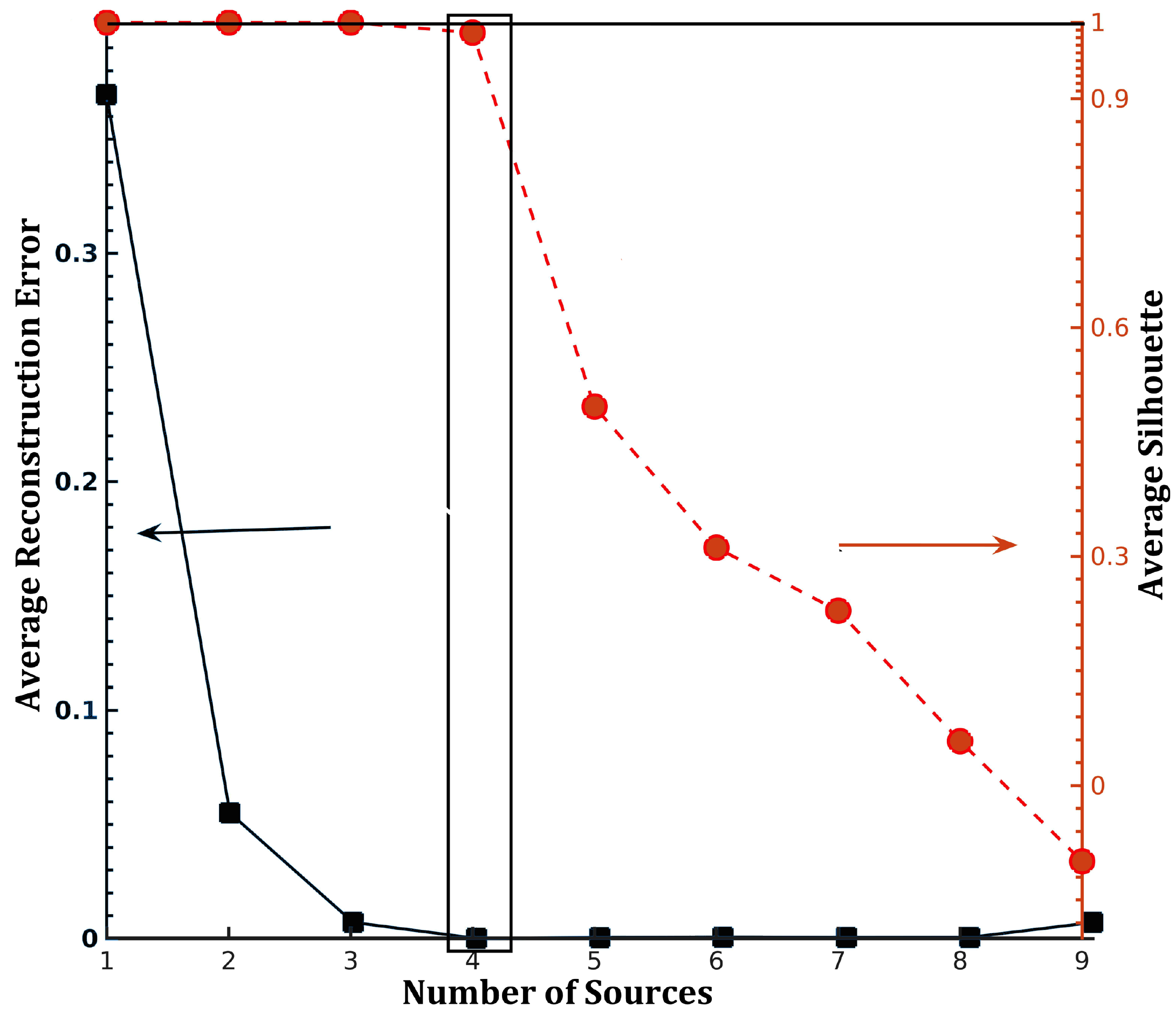}
\caption{Average reconstruction error (black) and the average Silhouette width (red) of the solutions for the case of nine detectors and four sources.
	The rectangle outlines the results related to the optimal number of sources.}
\label{fig:d9s4}
\end{figure}

The first setup (Fig.\ref{fig:maps}A) has four point sources and nine detectors.
In Fig.\ref{fig:d9s4}, we demonstrate the average reconstruction error and average Silhouette width obtained for different number of unknown sources $i$.
The results unambiguously point to the conclusion that there are four sources, $N_s =4$.
The average Silhouette width first slightly decreases as we move from one to three sources, while the reconstruction error remains relatively high; then the Silhouette width drops sharply as we go from four to five sources (Fig.\ref{fig:d9s4}).
With the increase in the number of sources from $4$ to $5$ and so on, we also observe that the reconstruction error remains almost the same, while the average Silhouette width of the clusters decreases.
The values of the $AIC$ function given in Table \ref{tbl:AIC} also confirm this conclusion.
The advection velocity estimated by the method is $u = 0.050214$ km/year, and the dispersion components are $D_x = 0.0050012$ km/year$^2$ and $D_y = 0.0012515$ km/year$^2$,
These estimates are in very good agreement with the actual values ($u = 0.005$ km/year, $D_x = 0.005$ km/year$^2$ and $D_y = 0.00125$ km/year$^2$).
The estimated source coordinates and strengths are given in Table \ref{tbl:resutls} and are also very accurate. 

The next three analyses are based on the second setup presented on Fig.\ref{fig:maps}B.

\begin{figure}
\centering\includegraphics[width=5.5in]{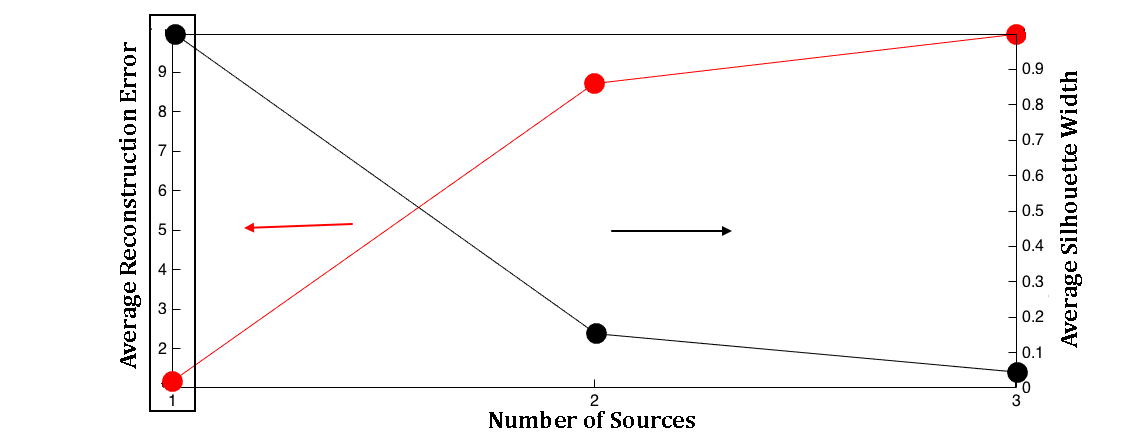}
\caption{Average reconstruction error (red) and the average Silhouette width (black) of the solutions for the case of three detectors and a single source.
	The rectangle outlines the results related to the optimal number of sources.}
\label{fig:d3s1}
\end{figure}

We start with only one point source, $S3$, with coordinates $(-0.2, 0.6)$ and strength $q=0.3$ mg/L, and three detectors: $D3, D2, D4$.
In Fig.\ref{fig:d3s1}, we demonstrate the average reconstruction error and the average Silhouette widths obtained at the end of the procedure for different number of sources $i$.
The method correctly selects only one source, and the values of the $AIC$ function given in Table \ref{tbl:AIC} are in agreement.
With the increase of the number of possible sources from $1$ to $2$ and then $3$, we observe that the reconstruction error grows, while the Silhouette width of the clusters decreases.
The methods yields for the advection velocity $u = 0.050125$ km/year, and for $D_x = 0.005002$ km/year$^2$ and $D_y = 0.0012475$ km/year$^2$.
The source parameters provided in Table \ref{tbl:resutls} also demonstrate a good agreement between the estimated and the actual parameter values.

Next, we use another combination of sources, $S1$ and $S2$, and detectors $D1$, $D2$, $D3$, and $D4$ (Fig.\ref{fig:maps}B).
Again, the algorithm results convincing identify the correct number of sources, $N_s = 2$ (Fig.\ref{fig:d3s1}).
The Silhouette widths first slightly decrease as we move from one to two sources in our reconstruction procedure, and then drop sharply as we go from two to three (Fig.\ref{fig:d3s1}).
Combined with the accompanying increase in the reconstruction error between two and three sources, this clearly points to two sources as the best estimate for $N_s$.
The values of the $AIC$ function (given in Table \ref{tbl:AIC}) support this conclusion.
The estimated parameters are also very good: $u = 0.05224$ km/year,  $D_x = 0.0050125$ km/year$^2$ and $D_y = 0.0012496$ km/year$^2$
(see Table \ref{tbl:resutls}).
\begin{figure}
\centering\includegraphics[width=5.5in]{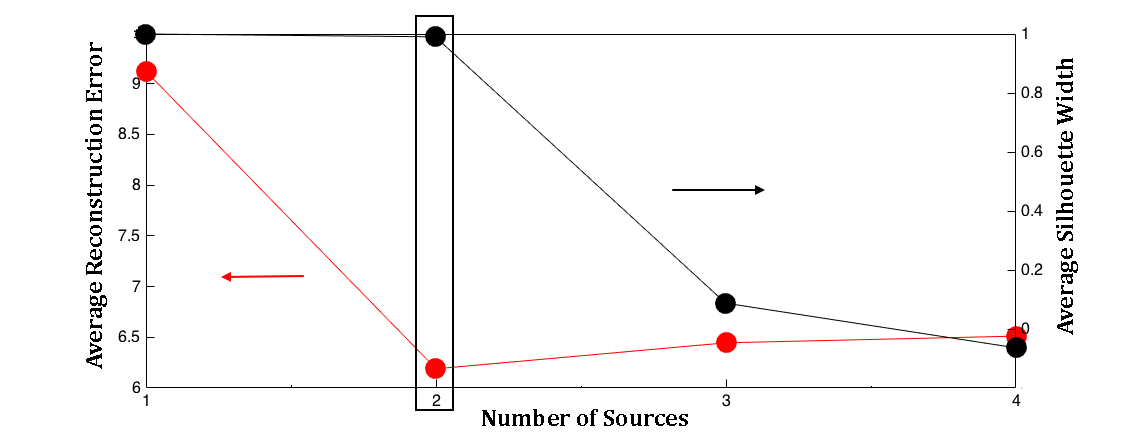}
\caption{Average reconstruction error (red) and the average Silhouette width (black) of the solutions for the case of four detectors and two sources.
	The rectangle outlines the results related to optimal number of sources.}
\label{fig:d4s2}
\end{figure}

\begin{figure}[H]
\centering\includegraphics[width=5.5in]{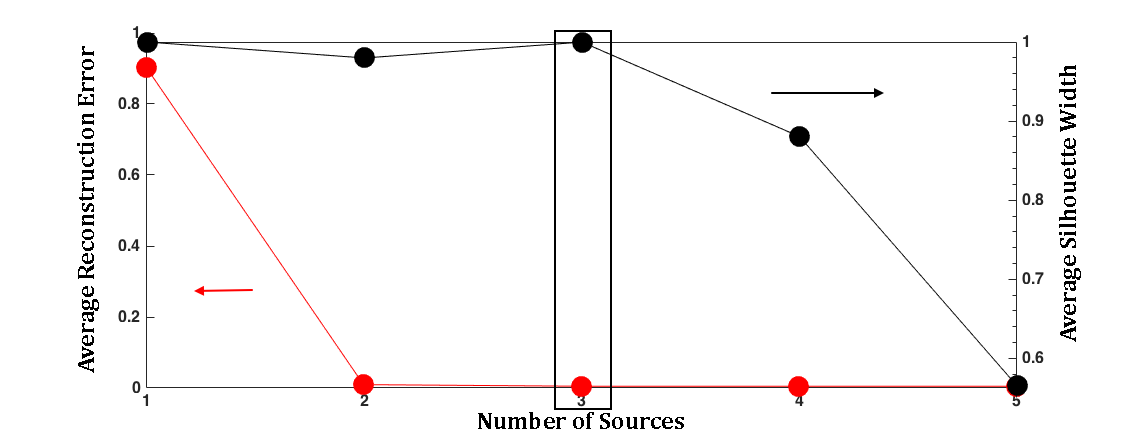}
\caption{Average reconstruction error (red) and the average Silhouette width (black) of the solutions for the case of five detectors and three sources.
	The rectangle outlines the results related to optimal number of sources.}
\label{fig:d5s3}
\end{figure}

In our last test case for instantaneous point-like sources, the signals are mixtures of all three sources, $S1$, $S2$, and $S4$, that are detected by each of the five detectors, forming a regular array Fig.\ref{fig:maps}B.
As it can be seen in Fig.\ref{fig:d5s3}, while the reconstruction error becomes quite small and flat for $N_s > 2$, the Silhouette width sharply drops for $i > 3$.
Thus, the most robust answer for the unknown number of sources $i$ is three.
The medium parameters are: $u = 0.051341$ km/year, $D_x = 0.005132$ km/year$^2$, $D_y = 0.0012512$ km/year$^2$;
Table \ref{tbl:resutls} lists the source parameters. 
The values of the $AIC$ function are given in Table \ref{tbl:AIC}.
Based on the results presented in Table \ref{tbl:AIC} and the figures in this section, the combination of Silhouette width cut-off with the $AIC$ criteria is an accurate method for estimating the unknown number of sources.
This is important observation considering that these metrics (Silhouette width and $AIC$) explore very different properties of the estimated solutions.
The Silhouette width focuses on the robustness and reproducibility of the solutions, while the $AIC$ score evaluates the solutions' parsimony.

The true and calculated parameters of the sources for all considered configurations are given in Table \ref{tbl:resutls}.
From the values listed in the table, it can be seen that the HNMF method is remarkably accurate, and the calculated coordinates are within $\sim$ 2\% of their true values. 

\begin{table}[H]
\centering
\caption{$AIC \times 10^{-3}$ estimates for a different unknown number of sources $i$ for all the test cases.
	The values of $i$ which are rejected because of bad clustering (Silhouette width $< 0.7$) are given in parentheses.
	Values of $i$ which exceed the number of detectors are denoted as not applicable ("na").}
\label{tbl:AIC}
\tiny{
	\begin{tabular}{|p{15mm}|c|c|c|c|c|c|c|c|c|}
		\hline 
		Case & \multicolumn{9}{|c|}{$i$ (unknown number of sources)} \\
		\cline{2-10}
		& 1 & 2 & 3 & 4 & 5 & 6 & 7 & 8 & 9\\
		\hline 
		\hline
		$1$ source \newline $3$ detectors & \bf{-1.262} & -0.714&  -0.7 & na & na & na & na & na & na \\
		\hline    
		\hline
		$2$ sources \newline $4$ detectors & -1.126& \bf{-1.236}& (-1.209)& (-1.192) & na & na & na & na & na \\
		\hline
		\hline
		$3$ sources \newline $5$ detectors & -0.978& -4.229& \bf{-4.47}& (-4.455)& (-4.438) & na & na & na & na \\
		\hline    
		\hline
		$4$ sources \newline $9$ detectors & 
		-5.431& -6.777& -8.229& \bf{-10.942}& (-10.794)& (-10.204)&( -9.865)& (-8.209)& (-8.024) \\
		\hline        
	\end{tabular}
}
\end{table} 
\begin{table}[H]
\centering
\caption{HNMF results presenting estimated model parameters for $1$, $2$, $3$ and $4$ sources.}
\label{tbl:resutls}
\begin{tabular}{|p{25mm}|c|c|c|c|c|c|c|}
	\hline 
	Case & Source &\multicolumn{2}{|c|}{$q$} & \multicolumn{2}{|c|}{$x$} & \multicolumn{2}{|c|}{$y$} \\
	& & \multicolumn{2}{|c|}{mg/L} & \multicolumn{2}{|c|}{km} & \multicolumn{2}{|c|}{km}\\
	\cline{3-8}
	& & true & est. & true & est. & true & est. \\
	\hline
	\hline
	$1$ source \newline $3$ detectors & \#1 & 0.3 & 0.299 & -0.2 &-0.198 & 0.6 & 0.599\\ 
	\hline 
	\hline
	$2$ sources & \#1 & 0.5 & 0.511 & -0.1& -0.100 &-0.2 & -0.200 \\
	\cline{2-8}
	$4$ detector & \#2 & 0.7 & 0.704 & -0.9 &-0.899 & -0.8 & -0.801\\
	\hline \hline
	$3$ sources & \#1 & 0.3 & 0.297 & -0.2 & -0.201 & 0.6 & 0.597\\ \cline{2-8}
	$5$ detectors & \#2 & 0.5 & 0.499 &-0.9 & -0.899 &-0.8 &-0.744\\
	\cline{2-8}
	& \#3 & 0.7 & 0.704 & -0.1 &-0.097 & -0.2& -0.199\\
	\hline \hline 
	$4$ sources & \#1 & 0.5 & 0.502 & -0.3 & -0.300 & 0.4 & 0.400\\ \cline{2-8}
	$9$ detectors & \#2 & 0.5 & 0.499 &-0.3 &-0.300&-0.4 &0.403\\ \cline{2-8}
	& \#3 & 0.5 & 0.500 &-0.3 & -0.301 & 0.65& 0.650\\
	\cline{2-8}
	& \#4 & 0.5 & 0.510 &-0.1 & -0.099 & -0.25& 0.249\\
	\hline 
\end{tabular}
\end{table}

\subsubsection{Time-dependent and spatially-extended sources and sources  in a bounded medium}
Here, we study three synthetic setups that geometrically are based on the combination of two sources, $S1$ and $S2$, and four detectors $D1$, $D2$, $D3$, and $D4$ (Fig.\ref{fig:maps}B). HNMF successfully identifies the unknown number of sources and their properties in three different cases: a) two extended sources, b) two sources with time dependence, and c) two point-like sources in medium with a reflecting boundary. In these three cases, we used the same coordinates and amplitudes for $S1$ and $S2$, and the same geometry of the detectors and the medium parameters as in the case of point-like sources in infinite medium already presented in the text. The combination of Silhouette width cut-off with the $AIC$ criterion produced accurate results for estimating the unknown number (two for each of these three cases) of sources. The behavior of the average reconstruction errors and the average Silhouette widths in each of these three cases is almost identical to that presented in Figure \ref{fig:d4s2} and these results are not shown here. The true and calculated parameters of the sources of the three cases are given in Table \ref{tbl:results2}. From the values listed in the table, it can be seen that the HNMF method is remarkably accurate in these specific cases and the errors are again within $\sim$ 2\% of their true values. The estimated values of the parameters of the medium, $ u_x, D_x$, and $D_y$,  are again within couple percents of their true values. 

First we consider the two sources, $S1$ and $S2$, to be instantaneous but to have finite sizes. Assuming constant source strength densities, and square shapes of the sources, the integral in (\ref{eq:Green_int}) can be done analytically:
\begin{eqnarray}
C({\bf x}, t) = \sum_{s=1}^{N_{s}} q_s\int_{x_s-d_s}^{x_s+d_s}\int_{y_s-d_s}^{y_s+d_s} d{x'} d{y'} G({\bf x}- {\bf x'},t - t_0) =
\nonumber\\
\sum_{s=1}^{N_{s}} \frac{q_s}{4} \left[\mathrm{erfc}\left(\frac{x -x_s+d_s-u_x t }{2 \sqrt{D_x t}}\right)-
\mathrm{erfc}\left(\frac{x -x_s-d_s-u_x t }{2 \sqrt{D_x t}}\right)\right]\times  \nonumber\\
\left[\mathrm{erfc}\left(\frac{y -y_s+d_s }{2 \sqrt{D_y t}}\right)-
\mathrm{erfc}\left(\frac{y -y_s-d_s}{2 \sqrt{D_y t}}\right)\right].     
\end{eqnarray}
Here $q_s$ is the source strength density, for a source centered at $(x_s, y_s)$ with size $2d_s\times 2d_s$ acting at $t=t_0$.
The position and extend of each source is specified by three unknown parameters ($x_s, y_s, d_s$), versus two for a point-like source. For the sizes of the squares for plumes $S1$ and $S2$ we used  0.2 and 0.1. HNMF estimated these to be 0.203 and 0.108, respectively. 

Next, we consider $S1$ and $S2$ to be point-like sources that work continuously since the initial time $t_0$. In this case the integrals in (\ref{eq:Green_int}), 
\begin{eqnarray}
C({\bf x}, t) = \sum_{s=1}^{N_{s}} q_s \int dt' G({\bf x}- {\bf x'},t - t'), 
\end{eqnarray}
has to be calculated numerically. Above, $G({\bf x}- {\bf x'},t - t')$ is the standard instantaneous Green's function given in (\ref{eq:Green}).

Finally, we consider $S1$ and $S2$ to be point-like sources in a 2D medium with a reflecting boundary.
The Green's function for (\ref{eq:pde}) is well-known \cite{wang2009analytical}, and can be written (for $0 \leqslant y \leqslant \infty$) as:
\begin{eqnarray}\label{eq:GreenBC}
G({\bf x},t) = \frac{1}{4 \pi \sqrt{D_x D_y} t} e^{-\frac{(x -x_s - u_x t)^2}{4 D_x t}}( e^{-\frac{(y - y_s)^2}{4 D_y t}} + e^{-\frac{(y + y_s)^2}{4 D_y t}}) ,
\end{eqnarray}
where $x$ and $y$ are the components of the vector ${\bf x}$ and $t>0$. $G({\bf x},t)$ satisfies boundary condition: $G_{|_{y->+\infty}} = 0$ and ${\partial G/\partial y}_{|_{y=0}} = 0$.
The reflecting boundary along the plane at $y=0$ in this case defined the top of the aquifer (domain) where the flow occurs.
Typically, the top of the aquifer is defined by either water-table or confining hydrogeologic strata.

\begin{table}[H]
\centering
\caption{HNMF results of the estimated model parameters for $S1$, $S2$ sources in the described three different cases described above.}
\label{tbl:results2}
\begin{tabular}{|p{25mm}|c|c|c|c|c|c|c|}
\hline 
Case & Source &\multicolumn{2}{|c|}{$q$} & \multicolumn{2}{|c|}{$x$} & \multicolumn{2}{|c|}{$y$} \\
& & \multicolumn{2}{|c|}{mg/L} & \multicolumn{2}{|c|}{km} & \multicolumn{2}{|c|}{km}\\
\cline{3-8}
& & true & est. & true & est. & true & est. \\
\hline 
\hline                    
Spatially-extended instantaneous sources & \#1 & 0.7 & 0.69 & -0.9 & -0.901 &-0.8 & -0.802\\
\cline{2-8}
& \#2 & 0.5 & 0.45 & -0.1 &-0.107 & -0.2 & -0.201\\
\hline 
Point continuous sources & \#1 & 0.7 & 0.7001 & -0.9 & -0.898 &-0.8 & -0.802\\
\cline{2-8}
& \#2 & 0.5 & 0.4999 & -0.1 &-0.0996 & -0.2 & -0.1998\\
\hline 
Point sources in a bounded media & \#1 & 0.7 & 0.704 & -0.9 & -0.9003 &-0.8 & -0.8002\\
\cline{2-8}
& \#2 & 0.5 & 0.4989 & -0.1 &-0.1009 & -0.2 & -0.2001\\
\hline 
\end{tabular}
\end{table}

\begin{figure}[H]
\centering
\includegraphics[width=6in]{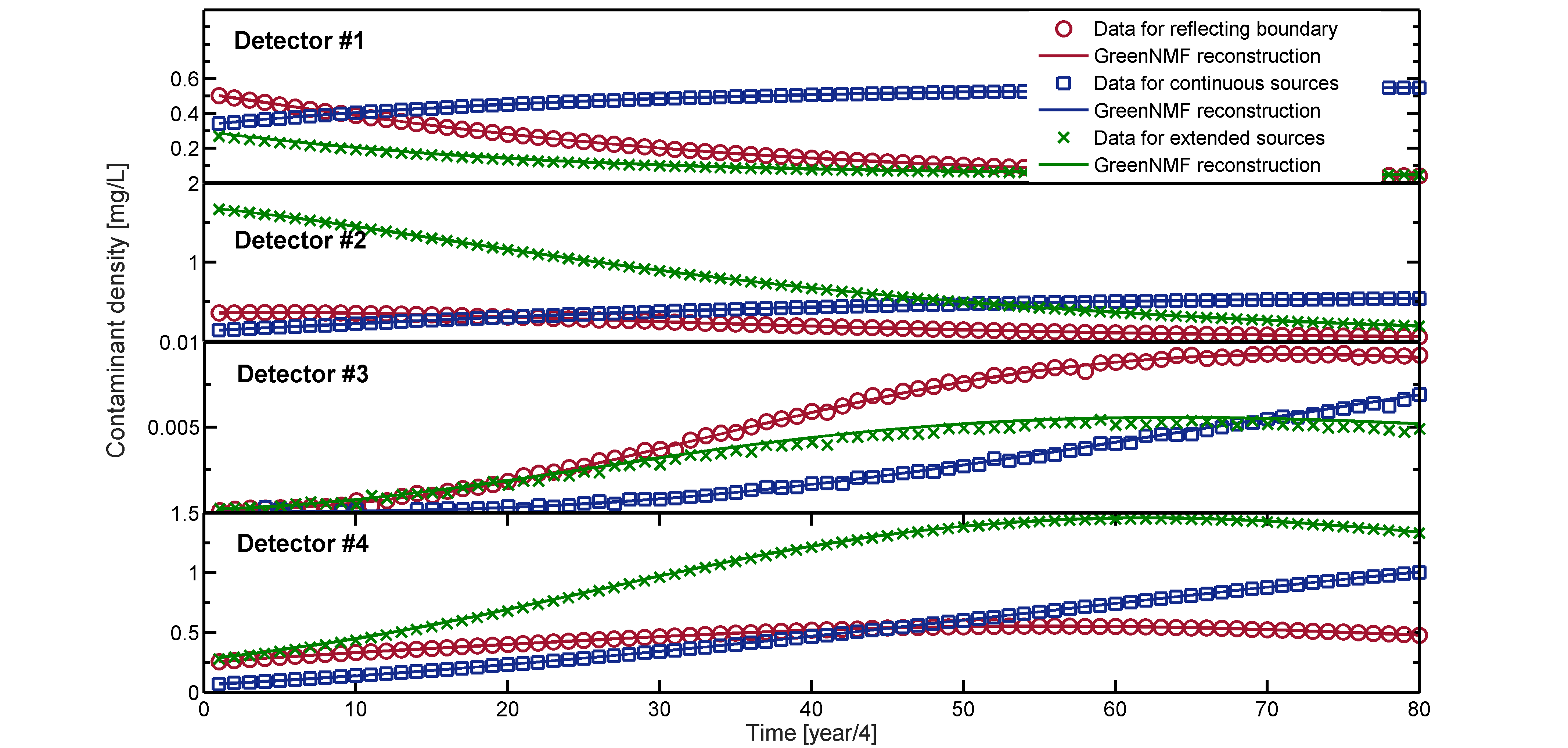}
\caption{Lines show the mixtures generated from the exact solutions at the detectors $D1$, $D2$, $D3$, $D4$ for the three cases with two sources and four detectors. The observations reconstructed by HNMF method are shown with different markers in the same colors as the data.}
\label{fig:source}
\end{figure}

In Figure \ref{fig:source} we present the generated mixtures and their reconstructions by HNMF for the three cases with two sources $S1$, $S2$ :  a) extended instantaneous sources, b) point-like continuous sources and c) point-like instantaneous sources in a bounded media. 

\section{Conclusions}
In this work we presented a new hybrid method, HNMF, that couples machine-learning algorithms with an inverse-analysis technique. The goal of the method is the identification of an unknown number of release sources of particles or energy based on spatiotemporal distributed set of observations.
HNMF uses the explicit form of the Green's function of the governing partial differential equation, combined with Non-negative Matrix Factorization, a custom clustering algorithm, and an iterative non-linear least-squares minimization procedure.
In the paper, HNMF is applied for identification of contaminant sources and model parameters.
To test the method, we generated several synthetic datasets, representing measurements recorded at a set of monitoring wells, and describing mixtures of unknown contaminant sources in an aquifer.
It is assumed that each source is releasing the same geochemical constituent, mixed in the aquifer, and the resultant mixture is detected at the observation points.
Using only the observations, HNMF correctly unmixed the contamination plumes from the mixtures observed in the monitoring wells.
Based on this unmixing, the method accurately identified the number and the locations of the sources, as well as the properties of the contaminant migration through the flow medium (advection velocity and dispersion coefficients).

To address the basic under-determinacy in the inverse problem  -- the unknown number of sources --   HNMF explores the plausible inverse solutions and seeks to narrow the set of possibilities by estimating the optimal number of contaminant source signals that robustly and accurately reconstruct the available data.
This allows us to estimate the number of contaminant sources, and determine their locations and strengths, as well as several important parameters of the medium such as dispersivity and advection velocity.
Future work will include the application of the HNMF to a real-world data as well as probabilistic analyses of the uncertainty associated with the solutions identified by HNMF.

Here we have concentrated on only one (albeit very important) application of HNMF  -- the problem of groundwater contamination.
However, there are numerous fields in which this method could be useful, since many diverse phenomena, such as heat flow, infectious disease transmission, population dynamics, spreading chemical/biochemical substances in the atmosphere, and many others, can be modeled by the advection-diffusion equation.
In general, the HNMF method presented here can be applied directly to any problem that is subject to the partial-differential advection-diffusion equation, where mixtures of an unknown number of physical sources are monitored at multiple locations.
More generally, the HNMF approach can be also applied to a whole set of different processes with distinct Green's functions; for example, Green's function representing anomalous (non-Fickian) dispersion \cite{OMalley2014a} or wave propagation in dispersive media. An open source Matlab implementation of HNMF method, that can be used for identification of a relatively small number of signals, can be found at: https://github.com/rNMF/HNMF. Finally, HNMF is a part of the submitted ``Source Identification by Non-negative Matrix Factorization Combined with Semi-Supervised Clustering,'' U.S. Provisional Patent App.No.62/381,486, filed by LANL on 30 August 2016.

\section{Acknowledgements}
This research was funded by the Environmental Programs Directorate of the Los Alamos National Laboratory.
In addition, Velimir V. Vesselinov was supported by the DiaMonD project (An Integrated Multifaceted Approach to Mathematics at the Interfaces of Data, Models, and Decisions, U.S. Department of Energy Office of Science, Grant \#11145687).
Velimir V. Vesselinov and Boian S. Alexandrov were also supported by LANL LDRD grant 20180060.

\end{document}